\documentclass[10pt,twocolumn,letterpaper]{article}

\usepackage[pagenumbers]{cvpr} 

\usepackage[dvipsnames]{xcolor}

\usepackage{times}
\usepackage{epsfig}
\usepackage{graphicx}
\usepackage{amsmath}
\usepackage{amssymb}
\usepackage{algorithm}
\usepackage{algorithmic}
\usepackage{multirow}
\usepackage{array}
\usepackage{booktabs}
\usepackage{makecell}
\usepackage{float}
\usepackage{diagbox}
\usepackage{changepage}
\usepackage{cuted} 
\usepackage{xcolor}
\usepackage{caption}
\usepackage{subcaption}
\usepackage{dsfont}
\usepackage{chngcntr}
\usepackage{color, colortbl}
\usepackage{makecell}
\usepackage{bbding}
\definecolor{LightCyan}{rgb}{0.91,0.91,0.98}
\definecolor{LightYellow}{rgb}{1.0, 1.0, 0.88}
\definecolor{magicmint}{rgb}{0.67, 0.94, 0.82}
\definecolor{lightmauve}{rgb}{0.86, 0.82, 1.0}
\definecolor{grannysmithapple}{rgb}{0.66, 0.89, 0.63}
\definecolor{isabelline}{rgb}{0.95,0.93,0.91}

\definecolor{cvprblue}{rgb}{0.21,0.49,0.74}
\usepackage[pagebackref,breaklinks,colorlinks,citecolor=cvprblue]{hyperref}

\def\paperID{4240} 
\def\confName{CVPR}
\def\confYear{2024}

\title{OpenSD: Unified Open-Vocabulary Segmentation and Detection}

\author{Shuai Li$^{1,2}$ \quad  Minghan Li$^{1,2}$ \quad   Pengfei Wang$^{1}$ \quad Lei Zhang$^{1,2}$\thanks{Corresponding author.} \\
{$^{1}$The Hong Kong Polytechnic University \qquad $^{2}$OPPO Research Institute}\\
}

\begin{document}
\maketitle
\begin{abstract}
Recently, a few open-vocabulary methods have been proposed by employing a unified architecture to tackle generic segmentation and detection tasks. However, their performance still lags behind the task-specific models due to the conflict between different tasks, and their open-vocabulary capability is limited due to the inadequate use of CLIP. 
To address these challenges, we present a universal transformer-based framework, abbreviated as OpenSD, which utilizes the same architecture and network parameters to handle open-vocabulary segmentation and detection tasks. First, we introduce a decoder decoupled learning strategy to alleviate the semantic conflict between thing and staff categories so that each individual task can be learned more effectively under the same framework. Second, to better leverage CLIP for end-to-end segmentation and detection, we propose dual classifiers to handle the in-vocabulary domain and out-of-vocabulary domain, respectively. The text encoder is further trained to be region-aware for both thing and stuff categories through decoupled prompt learning, enabling them to filter out duplicated and low-quality predictions, which is important to end-to-end segmentation and detection.
Extensive experiments are conducted on multiple datasets under various circumstances. The results demonstrate that OpenSD outperforms state-of-the-art open-vocabulary segmentation and detection methods in both closed- and open-vocabulary settings.  Code is available at: \href{https://github.com/strongwolf/OpenSD}{https://github.com/strongwolf/OpenSD}.  

\end{abstract}    
\section{Introduction}
\label{sec:intro}
Image segmentation and detection are two widely studied topics in computer vision, aiming to predict the masks and locations of objects in an image. With the rapid development of deep learning techniques~\cite{resnet,vgg,densenet,mobilenets,googlenet1,vit,swin}, tremendous progresses on segmentation and detection have been achieved during the past decade, and many powerful segmentation and detection models have been developed, including Mask RCNN~\cite{maskrcnn}, DETR~\cite{detr}, Mask2former~\cite{mask2former}, \etc. However, due to the laborious and costly annotations, the number of categories of existing segmentation and detection datasets is still limited to a few dozens or hundreds, which are too small to represent the vocabulary occurs in real world. This greatly restricts the generality of segmentation and detection models in practical applications.
\begin{figure}[tbp]
    \centering
    \includegraphics[width=0.45\textwidth]{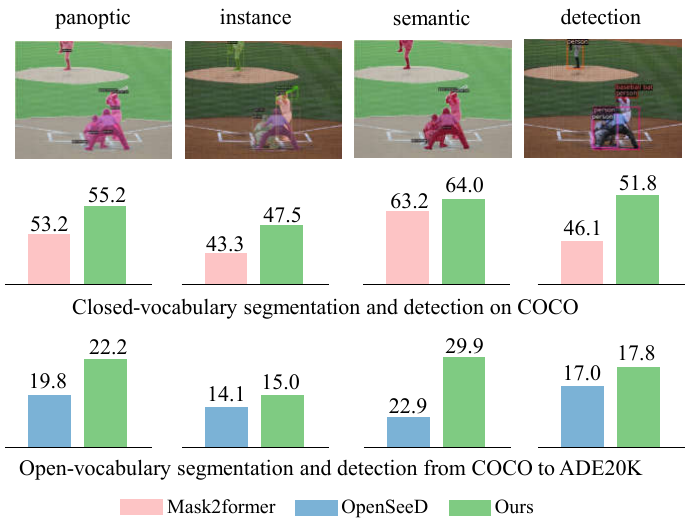}
    \caption{Our OpenSD achieves superior performance over previous methods (\eg, Mask2former \cite{mask2former} and OpenSeeD \cite{openseed}) in both closed- and open-vocabulary scenarios, including panoptic, instance and semantic segmentations, and object detection.}
    \label{fig_in_intro}
\end{figure}

To overcome the above mentioned limitations of closed-vocabulary segmentation and detection, open-vocabulary perception~\cite{lseg,openseg,f-vlm,simbaseline,vild}  has gained increasing research attention recently. Generally speaking, these approaches represent classifier weights as text embeddings of category names, which are extracted by a pre-trained text encoder in CLIP~\cite{clip}. By doing so, the model has the potential abilities to segment or detect objects of any categories that are described by natural languages. 

Most previous open-vocabulary segmentation and detection methods dedicate to designing specific models and learning strategies for different tasks. Though such task-specified models~\cite{lseg,openseg,f-vlm,simbaseline,vild} have advanced each individual task, they lack the flexibility to generalize to other tasks and the model needs to be trained from scratch when switching to a new task. To address this problem, some methods have been proposed to design a universal architecture to handle different segmentation tasks. For example, Mask2former~\cite{mask2former} can accomplish multiple segmentation tasks using one compact system; however, it needs to train a customized model for each task and lacks the open-vocabulary capability. Recent methods such as X-Decoder~\cite{x-decoder}, SEEM~\cite{seem} and OpenSeeD~\cite{seem} all utilize a similar architecture to Mask2former for generic segmentation by replacing the original classifier in Mask2former with the text encoder in CLIP. However, their open segmentation capability is limited since the text encoder is finetuned on the training dataset and the visual encoder of CLIP is not fully leveraged. FreeSeg~\cite{freeseg} employs a unified framework for different segmentation tasks, but it requires multiple inference passes for different tasks. In addition, its performance lags behind the individual task-specific models as the conflicts among different tasks are not carefully considered.

To overcome the above problems, we propose a unified open-vocabulary framework for effective segmentation and detection, abbreviated as OpenSD, which is built upon an encoder-decoder architecture and unifies the different segmentation and detection tasks into a two-stage paradigm. The first stage of OpenSD extracts class-agnostic object masks and boxes, and the second stage predicts classification scores based on these masks and boxes. To make OpenSD succeed, there are two key challenges to be addressed, \ie, how to mitigate the conflicts among different tasks, and how to effectively utilize CLIP. 

First, the different segmentation and detection tasks have potential conflicts when unifying them into the same framework.  Instance segmentation and detection require recognizing objects of foreground-thing categories, while semantic segmentation focuses on both foreground-stuff and background-stuff categories. Simply using a shared decoder and query set for all tasks can result in unsatisfactory performances due to the semantic conflict between thing and stuff categories. Considering this, we propose a decoder decoupled learning strategy to reduce the interaction between thing and stuff queries in the self- and cross-attention layers in the decoder. In this way, the learning efficiency of each individual task can be enhanced under the same framework.

Second, how to effectively utilize CLIP for end-to-end segmentation and detection needs in-depth investigation. Though CLIP is the key to the open-vocabulary ability of segmentation and detection models, previous studies show that CLIP is not a cure-all for all domains and categories and it struggles to balance between the generalization ability in the out-of-vocabulary domain and the performance in the in-vocabulary domain (\eg, training domain). 
To mitigate this issue, we propose dual classifiers, where a pre-trained CLIP text encoder is employed to predict classification scores for both query embeddings (in-vocabulary domain) and CLIP embeddings (out-of-vocabulary domain).
In addition, since CLIP is pre-trained on image-text data, it has no discrimination ability for low-quality or duplicated predictions generated by the first stage. To overcome this problem, we further propose to adapt the text encoder to be region-aware for both thing and stuff categories by decoupled prompt learning so that duplicated and low-quality predictions can be effectively removed. 

Thanks to our effective and careful design, OpenSD achieves superior performance over competing methods on various tasks, including semantic segmentation, instance segmentation, panoptic segmentation and object detection in both open-vocabulary and closed-vocabulary settings, as shown in Fig.~\ref{fig_in_intro}. It provides a powerful framework for unifying image segmentation and detection tasks with open-vocabulary capabilities. 

\begin{figure*}[tbp]
    \centering
    \includegraphics[width=0.98\textwidth]{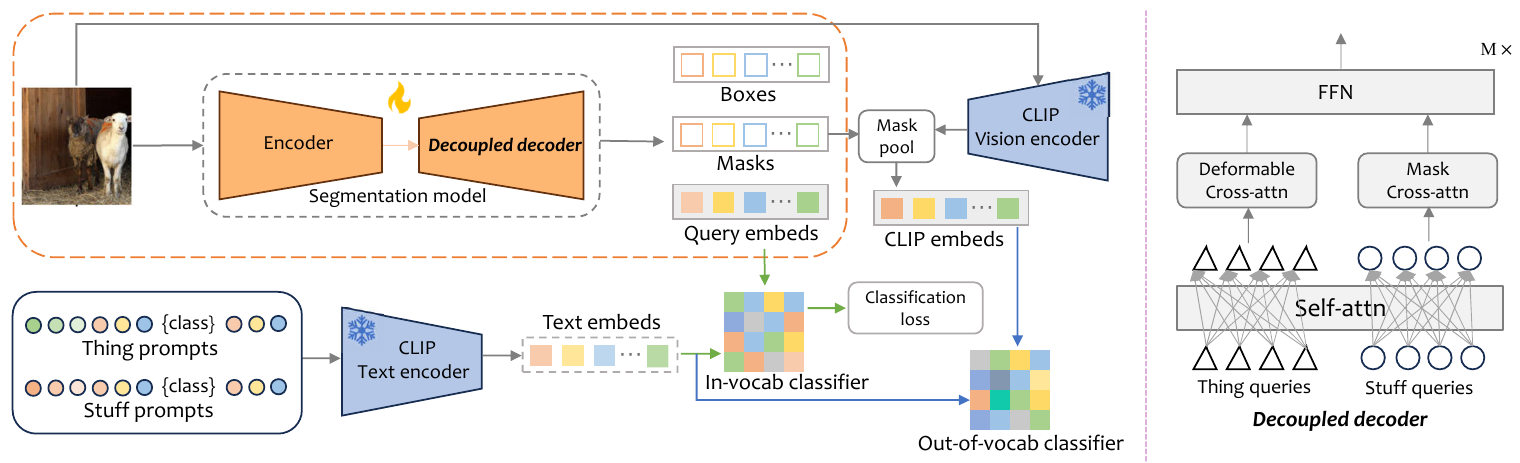}
    \caption{Overview of our OpenSD framework. In the first stage (highlighted in the orange rounded box), we utilize a customized segmentation model to generate several query embeddings, bounding boxes and masks. In the second stage, we employ dual classifiers, including an in-vocabulary classifier (green branch) and an out-of-vocabulary classifier (blue branch), to predict classification scores with respect to query embeddings and CLIP embeddings, respectively. To make text embeddings region-aware, we design decoupled prompts for thing and stuff categories. The in-vocabulary classifier is supervised by the one-to-one label assignment loss and the out-of-vocabulary classifier is only used during inference. Our proposed decoupled decoder learning strategy is shown on the right part of the figure, where query decoupling and attention decoupling are used to mitigate the semantic conflict between different tasks. }
    \label{framework}
\end{figure*}

\section{Related Work}
\label{sec:related_work}
Object detection and segmentation are two fundamental computer vision tasks~\cite{dpm,survey-deepsem,survey-det,survey-sem}, while segmentation can be further divided into semantic segmentation~\cite{deeplab,fcn,sem-oc,sem-pcp}, instance segmentation~\cite{yolact,condinst,solo,maskrcnn,queryinst,Li_2021_CVPR} and panoptic segmentation~\cite{pano-deeplab,pano-fcn,pano-fpn,pano-seg,pano-upsnet} with respect to different semantics. Recently, transformer-based methods~\cite{attention} have become prevalent in the development of both segmentation~\cite{mask2former,maskdino,oneformer,mdqe} and detection~\cite{detr,deformdetr}. However, most of these methods are constrained to a limited size of closed vocabulary, and there is a high demand for open-vocabulary segmentation and detection methods.

\textbf{Open Vocabulary Segmentation and Detection.}
Previous studies~\cite{simbaseline, rao2022denseclip, odise, groupvit} usually address the challenge of open-vocabulary segmentation and detection by utilizing label embeddings.
Many investigations are focused on pretraining models tailored for either detection or segmentation using large-scale image-text paired data or fully annotated data.  For instance, GLIP~\cite{glip}, Grounding DINO~\cite{groundingdino} and RegionCLIP~\cite{regionclip} generate pseudo boxes and labels on image-text data and employ region-text contrastive learning for detection. In the case of segmentation, OpenSeg~\cite{openseg} predicts class-agnostic masks using large-scale captioning data and learns visual-semantic alignment based on the predicted masks. Detic~\cite{detic} utilizes image classification data to expand the vocabulary for detection. However, pretraining on image-text data can be computationally intensive, and the open-vocabulary capability can be affected by the reliability of the pseudo labels.
X-decoder~\cite{x-decoder}, SEEM~\cite{seem}, and OpenSeeD~\cite{openseed} all employ an encoder-decoder architecture and fine-tune the models via supervised training on COCO and Object365 datasets. LSeg~\cite{lseg} aligns the pixel embeddings from the image encoder with text embeddings through contrastive learning to improve semantic segmentation. These methods excel in closed domains but face challenges when encountering unseen categories.

Another line of research directly utilizes the open visual recognition capabilities of vision-language models (VLMs) such as CLIP~\cite{clip}. ViLD~\cite{vild} distills the feature representation of CLIP into the detector. F-VLM~\cite{f-vlm} presents an open-vocabulary detection approach based on a frozen VLM. In segmentation, using the frozen CLIP as the image visual encoder has also been explored. MaskCLIP~\cite{maskclip} introduces a Relative Mask Attention module to combine mask tokens with the CLIP backbone. SAN~\cite{san} employs an adapter on the CLIP image encoder to predict mask proposals and attention bias. OVSeg~\cite{ovseg} fine-tunes CLIP using a collection of mask image regions and the corresponding text descriptions. While these methods benefit from the open-vocabulary recognition ability of VLMs, their performance may become inferior for closed vocabularies.

\textbf{Unified Segmentation and Detection.}
While architectures for semantic segmentation, instance segmentation, panoptic segmentation and object detection can vary significantly, some methods~\cite{oneformer, k-net, mask2former} have been proposed to unify these different tasks using the same architecture. One notable example is Mask2former~\cite{mask2former}. which employs a set of learnable queries that interact with the image encoder through the decoder, and uses the refined query features to predict category labels and masks. Mask2former is designed for closed vocabulary scenarios. Based on Mask2former, methods~\cite{freeseg, seem, x-decoder, uninext, openseed, fcclip} have been recently proposed to unify segmentation and detection in an open vocabulary context. However, many of these methods simply replace the classifier in Mask2former with label embeddings, without considering the potential conflicts between different tasks. In this work, we aim to design a unified learning framework to enhance the performance of various segmentation and detection tasks in both closed and open vocabulary scenarios.
\section{Method}
\label{sec:method}
We present a unified framework, namely OpenSD, to effectively tackle various open-vocabulary segmentation and detection tasks. OpenSD adopts a two-stage pipeline, as illustrated in Fig.~\ref{framework}. The first stage (indicated by the orange box) utilizes a customized segmentation model to generate several query embeddings, bounding boxes and masks. The second stage employs CLIP to predict classification scores based on the outputs of the first stage. In Sec.~\ref{sec3.1}, we describe a vanilla open-vocabulary segmentation method based on Mask2former \cite{mask2former}, and in Sec.~\ref{sec3.2} and Sec.~\ref{sec3.3}, we introduce in detail our elegant designs to elevate the vanilla pipeline into the powerful OpenSD framework.

\subsection{Mask2former with Label Embeddings}
\label{sec3.1}
Mask2former \cite{mask2former} is a strong closed-vocabulary segmentation model, which integrates semantic, instance, and panoptic segmentations using an encoder-decoder architecture. It incorporates multiple learnable object queries to represent both thing and stuff categories. The encoder extracts image features from the input image, while the decoder progressively updates the query embeddings through a series of decoder layers. Each decoder layer consists of a cross-attention layer, a self-attention layer, and an MLP. The updated query embeddings are finally utilized to convole with the image features to predict binary masks. Additionally, an MLP is applied after the query embeddings to predict classification scores for pre-defined categories. For detection purposes, an additional MLP can be included to predict the coordinates of bounding boxes.

To adapt Mask2former for open-vocabulary segmentation and detection, a straightforward solution is to replace the MLP classifier with label embeddings extracted from the pre-trained text encoder in CLIP, and then train on detection and segmentation datasets to align the query features to the embeddings of the text encoder using seen categories. As a result, the model may gain a certain level of open-vocabulary capability on unseen categories. Such a solution serves as a baseline method in this paper.

\begin{figure}[tbp]
    \centering
    \includegraphics[width=0.45\textwidth]{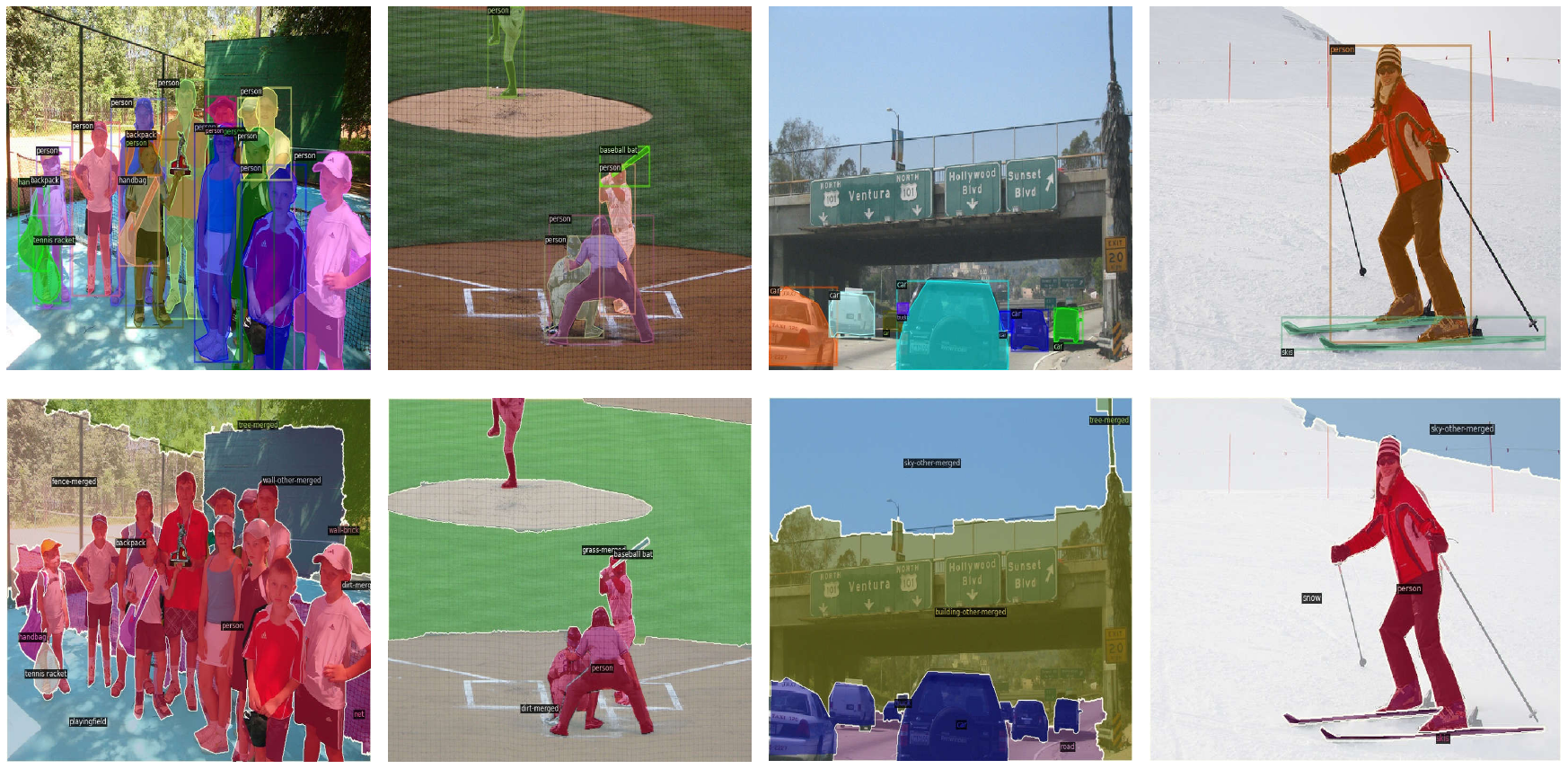}
    \caption{GT masks for thing categories in instance segmentation (top) and stuff categories in semantic segmentation (bottom). }
    \vspace{-1mm}
    \label{thing_stuff_gt_masks}
    \vspace{-3mm}
\end{figure}

\subsection{Decoupled Decoder Learning}
\label{sec3.2}
The learning efficiency of Mask2former is not very high when using the same network parameters for all tasks because it does not carefully consider the discrepancies between different tasks, where there are two types of categories: thing and stuff. Thing category represents countable objects that need to be distinguished during inference, while stuff category is non-countable and indistinguishable. The discrepancies primarily arise from the distinct characteristics of different categories.
Foreground categories have varied semantics across different tasks. As shown in Fig.~\ref{thing_stuff_gt_masks}, `person' is considered as a thing category in detection and instance segmentation, but it is categorized as a stuff category in semantic segmentation. Also, thing and stuff categories may exhibit significantly different distributions in terms of area, position and shape. For example,
stuff categories like `sky', `building' and `grass' in Fig.~\ref{thing_stuff_gt_masks} often occupy a much larger portion of the image than thing categories.
Utilizing the same object queries and decoder to represent all categories is sub-optimal and can significantly degrade the performance of each individual task. To address this issue, we propose a decoupled decoder learning strategy, including query decoupling and attention decoupling, to maximize the learning efficiency of each task.

\textbf{Query Decoupling.} The primary purpose of object queries is to learn the possible positions where objects are likely to appear within an image. However, due to the significant variation in spatial distributions between thing and stuff categories, it is challenging to learn the optimal query embeddings for both categories using the same object queries. To address this challenge, we propose a decoupled learning approach for object queries by separating them into thing queries ($Q_t$) and stuff queries ($Q_s$), which are independently decoded through the decoder.

Specifically, the thing queries and stuff queries are passed through the decoder, resulting in separate outputs: $(P_t^c, P_t^m, P_t^b)$, where $P^c$, $P^m$ and $P^b$ represent the predictions of classification scores, masks and boxes, respectively. Additionally, we divide the ground truths into two sets: thing ground truths (T-GTs) and stuff ground truths (S-GTs). T-GTs encompass the foreground categories in detection, instance segmentation and panoptic segmentation, while S-GTs include both the foreground and background categories in semantic segmentation, and backgroud categories in panoptic segmentation. The outputs of the two sets are independently supervised by their respective GTs using the Hungarian Matching algorithm.

\textbf{Attention Decoupling.} To address the challenges posed by the different spatial distributions and semantic attributes of thing and stuff categories, we propose the decoupled cross-attention layers, which use independent branches to handle thing and stuff queries separately. Furthermore, we employ different types of cross-attention operations for these two layers to cater to their specific characteristics.

As shown in the right part of Fig.~\ref{framework}, for the stuff cross-attention layer, we adopt the mask attention mechanism~\cite{mask2former} to constrain the cross-attention within the entity region of the predicted mask. This mask attention serves as a global attention mechanism and is well-suited for stuff queries, as stuff entities often exhibit large sizes, irregular shapes and disconnected parts.
On the other hand, for the thing cross-attention layer, we adopt the deformable attention~\cite{deformdetr}, which aggregates discrete feature pixels from the feature map, to update query features. We found that the use of local features extracted by deformable attention is sufficient to update thing query features.
By employing decoupled cross-attention layers with distinct attention mechanisms, we can reduce the ambiguity in parameter learning and improve the model's ability to handle the diverse characteristics of thing and stuff categories.

\begin{figure}[tbp]
    \centering
    \includegraphics[width=0.45\textwidth]{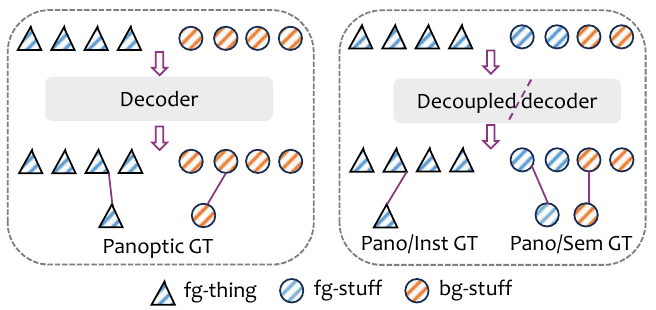}
    \caption{Comparison between OpenSeeD \cite{openseed} (left) and our OpenSD (right) in handling queries.
    }
    \label{compare_with_openseed}
\end{figure}

\textbf{Remarks.} 
Fig.~\ref{compare_with_openseed} compares our approach and a recent method OpenSeeD \cite{openseed}, which divides queries into foreground and background ones, supervised by panoptic annotations. In contrast, we divide the queries into thing queries and stuff queries, supervised by both panoptic, instance and semantic annotations. Our query division strategy covers a wider range of tasks during training and it is more compatible with the inference process. Furthermore, OpenSeeD uses a single decoder, while our method incorporates attention decoupling in the decoder to effectively mitigate the conflicts between different tasks.

\subsection{Region-aware Dual Classifiers}
\label{sec3.3}
We propose dual classifiers, which consist of an in-vocabulary classifier and an out-of-vocabulary classifier, to leverage query embeddings and CLIP embeddings for recognition. Query embeddings trained from supervised datasets are advantageous in predicting closed categories, while CLIP embeddings extracted from the CLIP visual encoder excel at open categories. A proper utilization of them is crucial for achieving excellent performance in both in-vocabulary and out-of-vocabulary domains.

The in-vocabulary classifier calculates a classification score for each object query by:
\begin{equation}
\begin{aligned}
P_{in}^i = sigmoid(1/T \times cosine(E_{text}(C^i), Q_{emb})),
\end{aligned}
\label{eq1}
\end{equation}
where $C^i$ is the prompt of the $i^{th}$ class name, $T$ is a temperature to amplify the logit,  $E_{text}$ is the pre-trained text encoder from CLIP, $Q_{emb}$ is a thing/stuff query embedding generated by the decoder.

Similarly, the predicted class probability by the out-of-vocabulary classifier is computed by:
\begin{equation}
\begin{aligned}
P_{out}^i = sigmoid(1/T \times cosine(E_{text}(C^i), \text{CLIP}_{emb})),
\end{aligned}
\label{eq2}
\end{equation}
where $\text{CLIP}_{emb}$ is the CLIP embeddings pooled from its vision encoder using the mask predictions $P^m$. $P_{in}^i$ and $P_{out}^i$ are then ensembled to get the final classification score.

For end-to-end segmentation and detection, classification scores are used to filter out low-quality and duplicated predictions. However, the original text encoder in CLIP lacks this ability as it is pre-trained for image-text alignment. To solve this problem, we propose to learn prompts for the text encoder $E_{text}$ to make it region-aware for both thing and stuff categories.

In particular, the original text encoder typically employs a fixed prompt template such as `{a photo of c', where c is the category name. For end-to-end segmentation and detection, a more suitable prompt could be something like `{a unique bounding box/mask of category c with high quality'. To encourage the text encoders to acquire such knowledge, we use `{$\star$ $\star$ $\star$ c $\star$ $\star$', where $\star$ represents a learnable vector, to replace the fixed prompts in Eq.~\ref{eq1} and Eq.~\ref{eq2}. To reduce the conflict between thing and stuff categories, we further propose decoupled prompts which contain category-specific prompts and category-shared prompts. Specifically, the thing category prompts are defined as `$\circ$ $\circ$ $\star$ $\star$ $\star$ c $\star$ $\star$' and the stuff category prompts are defined as `$\diamond$ $\diamond$ $\star$ $\star$ $\star$ c $\star$ $\star$', where $\circ$ is the thing-specific prompt, $\diamond$ is the stuff-specific prompt and $\star$ is the prompt shared by thing and stuff categories.
The learnable prompts of the text encoder are optimized on seen categories during training using the following losses:
\begin{equation}
-t^i \times \log(P_{in}^i) - (1 - t^i) \times (1 - \log(P_{in}^i)), 
\label{eq3}
\end{equation}
where $t^i$ is a discrete label indicating whether the current query is a positive or negative sample for category $i$. The value of $t^i$ for thing and stuff categories is determined by the Hungarian Matching results between the predictions of the decoder and the corresponding GTs.

\section{Experiments}
\begin{table*}[!t]
\centering
\scalebox{0.98}{
\setlength{\tabcolsep}{0.5mm}
\begin{tabular}{llc|cccc|cccc|ccc}
\Xhline{0.8pt}
\rowcolor{LightCyan}
\multicolumn{3}{c|}{Datasets} & \multicolumn{4}{c|}{COCO (closed)} &  \multicolumn{4}{c|}{ADE (open)} &  \multicolumn{3}{c}{Cityscapes (open)}  \\
\hline\hline
Method & Training Data & Backbone & PQ & mask AP & box AP & mIoU & PQ & mask AP & box AP & mIoU & PQ & mask AP & mIoU  \\
\hline
LSeg~\cite{lseg} & COCO & ViT-B & - & - & - & -& -& -& -& 18.0 & - & - & -  \\
OpenSeg~\cite{openseg} & COCO & ViT-B & - & - & - & -& -& -& -& 21.1 & - & - & -  \\
\hline 
Mask2former~\cite{mask2former} & COCO & R50 & 51.9 & 41.7 & 44.5 & 61.7 & - & - & - & - & - & - & - \\
OPSNet~\cite{opsnet} & COCO+IN & R50 & 52.4 & - & - & - & 17.7 & - & - & - & 37.8 & - & - \\
MaskCLIP~\cite{maskclip} & COCO & R50 & 30.9 & -& -& -& 15.1 & 6.0 & - & 23.7 & - & - & - \\
HIPIE~\cite{hipie} & COCO+O365 & R50 & 52.7 & \textbf{45.9} & \textbf{53.9} & 59.5 & 18.4 & 13.0 & 16.2 & 26.8 &  &  &  \\
\rowcolor{isabelline}
\textit{OpenSD} & COCO & R50 & 53.2 & 45.2 & 48.8 & 62 & 19.4 & 10.7 & 11.4 & 27.2 & 37.8 & 26.6 & \textbf{50.2} \\
\rowcolor{isabelline}
\textit{OpenSD} & COCO+O365 & R50 & \textbf{54} & 45.7 & 49.7 & \textbf{62} & \textbf{23} & \textbf{14.3} & \textbf{16.8} & \textbf{29.3} & \textbf{38.6} & 25.8 & 50.1 \\
\hline
Mask2former~\cite{mask2former} & COCO & Swin-T & 53.2 & 43.3 & 46.1 & 63.2 & - & - & - & - & - & - & - \\
X-Decoder~\cite{x-decoder} & COCO+ITP & Swin-T & 52.6 & 41.3 & 43.6 & 62.4 & 18.8 & 9.8 & - & 25.0 & 37.2 & 16.0 & 47.3 \\
SEEM~\cite{seem} & COCO & Swin-T & 50.6 & 39.5 & - & 61.2 & - & - & -& - & - & - & - \\
OpenSeeD~\cite{openseed} & COCO & Swin-T & 54.0 & 45.6 & 49.0 & 62.1 & 16.0 & 9.9 & 10.7 & 18.6 & 35.7 & 23.7 & 45.8 \\
\rowcolor{isabelline}
\textit{OpenSD} & COCO & Swin-T & 55.2 & 47.5 & 51.8 & 64 & 19.6 & 11.1 & 12.2 & 28.7 & 38.2 & \textbf{27.5} & \textbf{51.0}  \\
OpenSeeD~\cite{openseed} & COCO+O365 & Swin-T & 55.2 & 47.3 & 51.9 & 63.7 & 19.8 & 14.1 & 17.0 & 22.9 & 37.3 & 26.2 & 46.1 \\
\rowcolor{isabelline}
\textit{OpenSD} & COCO+O365 & Swin-T & \textbf{55.7} & \textbf{48} & \textbf{53.0} & \textbf{64.8} & \textbf{22.2} & \textbf{15.0} & \textbf{17.8} & \textbf{29.9} & \textbf{39.0} & 27.1 & 51.0  \\
\hline
ODISE~\cite{odise} &  COCO & ViT-H & 45.6 & 38.4& -& 52.4& 23.5 & 13.9 & - & 28.7 & - & - & -\\
\hline
Mask2former~\cite{mask2former} & COCO & Swin-L & 57.6 & 48.5 & 52.2 & 67.4 & - & - & - & - & - & - & - \\
X-Decoder~\cite{x-decoder} & COCO+ITP & Swin-L & 56.9 & 46.7 & - & 67.5 & 21.8 & 13.1 & - & 29.6 & 38.1 & 24.9 & 52.0 \\
OpenSeeD & COCO & Swin-L & 58.0 & 49.7 & 54.8 & 67.6 & 17.1 & 10.3 & 11.5 & 19.8 & 37.7 & 26.6 & 47.6 \\
\rowcolor{isabelline}
\textit{OpenSD} & COCO & Swin-L & \textbf{58.8} & \textbf{50.9} & \textbf{56.7} & \textbf{68.3} & \textbf{23.1} & \textbf{13.6} & \textbf{15.3} & \textbf{30.8} & \textbf{39.6} & \textbf{29.9} &  \textbf{52.2} \\
\Xhline{0.8pt}
\end{tabular}
}
\caption{Open-vocabulary segmentation and detection from COCO to ADE and Cityscapes. `IN' indicates ImageNet. `O365' indicates Object365. `ITP' indicates image-text pair data. The model is trained on COCO (or jointly with Object365) and tested on COCO (closed-vocabulary setting), and ADE and Cityscapes (open-vocabulary setting). `-' means that the result is not reported or the model does not have the ability for the specific task. \textit{It is worth noting that due to the significant training burden, we train our models on Object365 with much fewer iterations compared to the competiting methods.} In addition, since the training details and model parameters of OpenSeeD on Object365 with Swin-Large backbone are not publicly available, we re-train OpenSeeD on the COCO dataset using the official code to ensure a fair comparison.
} 
\label{coco_to_ade}
\end{table*}

\subsection{Implementation Details}
\textbf{Architecture.} We adopt Mask2former \cite{mask2former} as the segmentation model in our framework (see Fig.~\ref{framework}) by replacing the original decoder with our proposed decoupled decoder. By default, Swin-T pre-trained on ImageNet is used as the vision backbone of the encoder. The decoder consists of nine decoupled decoder layers to update the thing and stuff query features. Following \cite{openseed}, we set the number of thing and stuff object queries to 300 and 100, respectively. The pre-trained ConvNeXt-Base CLIP backbone and the text-encoder of CLIP from OpenCLIP~\cite{openclip} are employed.

\textbf{Datasets.} 
Our training and testing datasets mainly include COCO~\cite{coco}, ADE20K~\cite{ade}, and Object365~\cite{objects365}. COCO is a multi-task benchmark, including 118K training images and 5K validation images spanning 133 categories. ADE20K comprises 20K training images and 2K validation images, encompassing 150 categories. Object365 is an object detection dataset, with 660K and 1,700K 
images for versions 1 and 2, respectively.

\textbf{Evaluation Metrics.} 
We evaluate OpenSD in both closed-vocabulary and open-vocabulary settings. First, we train a model on the COCO training set and evaluate it on COCO test set, and ADE20K and Cityscapes~\cite{cityscapes} datasets, covering various segmentation and detection tasks.
For panoptic segmentation, we employ the Panoptic Quality (PQ) metric to evaluate the overall segmentation quality.
For semantic segmentation, we utilize the mean of Intersection over Union (mIoU) metric to measure the classification accuracy of different semantic categories.
For instance segmentation and detection tasks, we report the mean Average Precision (mAP) metric, which assesses the precision and recall of detected and localized objects.
In addition to training a model on COCO, we train a model on ADE20K and evaluate its performance on the COCO dataset. This allows us to assess how well the model can generalize across different datasets.
To further demonstrate the open-vocabulary performance of OpenSD, we also train an enhanced version of it by training on both COCO and Object365 datasets.

\textbf{Implementation Details.} For fair comparison, we adhere to the same training recipe of Mask2former \cite{mask2former} without any additional special designs. We employ the AdamW optimizer and train the model for 50 epochs.
The initial learning rate is set to 0.0001, which is decayed  at the 0.9 and 0.95 fractions of the total training epochs by a factor of 10. The number of category-specific and category-shared prompt vectors are set to 16 and 8, respectively.

\textbf{Inference Strategy.}
In testing, we resize the shorter side and longer side of the input image to min\_size and a maximum of max\_size. On COCO, the min\_size is 800 and the max\_size is 1,333. On ADE, the min\_size and max\_size are 640 and 2,560, respectively. On Cityscapes, the min\_size and max\_size are 1,024 and 2,048, respectively.

As shown in Fig.~\ref{supp-fig-inference}, we combine the outputs of the two decoders to obtain the final results for different tasks. The panoptic segmentation result is obtained by combining the foreground-query from the thing outputs and the background-query from the stuff outputs. The instance segmentation and object detection results are directly derived from the thing outputs. The semantic segmentation result is obtained by combining the foreground-query and background-query from the stuff outputs. The classification scores for each query are obtained by ensembling the scores from the in-classifier and out-of-classifier as follows:

\begin{equation}
P^{i}= 
    \begin{cases}
    (P_{in}^{i}) ^ {(1-\alpha)} \times (P_{out}^{i}) ^ {\alpha}, &  \textit{if  } \; i \in C_{train}, \\
    (P_{in}^{i}) ^ {(1-\beta)} \times (P_{out}^{i}) ^ {\beta},  & otherwise,
    \end{cases}
\label{supp-ensemble}
\end{equation}
where $P_{in}^{i}$ and $P_{out}^{i}$ are the classification scores of in-vocabulary classifier and out-of-vocabulary classifier with respect to category $i$, respectively. $\alpha$ and $\beta$ balance the contribution of the two classifiers for seen categories and unseen categories.

\begin{figure}[tbp]
    \centering
    \includegraphics[width=0.4\textwidth]{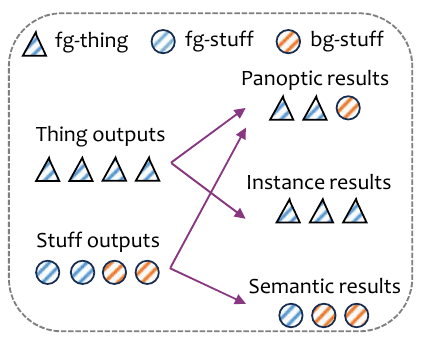}
    \caption{The assembling strategy for different tasks.}
    \label{supp-fig-inference}
\end{figure}

\begin{table*}[!t]
\centering
\scalebox{0.98}{
\setlength{\tabcolsep}{1.5mm}
\begin{tabular}{llc|cccc|cccc}
\Xhline{0.8pt}
\rowcolor{LightCyan}
\multicolumn{3}{c|}{Datasets} & \multicolumn{4}{c|}{ADE (closed)} &  \multicolumn{4}{c}{COCO (open)}   \\
\hline\hline
Method & Training Data & Backbone & PQ & mask AP & box AP & mIoU & PQ & mask AP & box AP & mIoU  \\
\hline
Mask2former~\cite{mask2former} & ADE & R50 & 39.7 & 26.4 & 28.8 & 47.7 & - & - & - & - \\
ZSSeg~\cite{simbaseline} & ADE & R101 & - & - & - & - & 11.2 & 4.3 & - & 17.7 \\
FreeSeg~\cite{freeseg} & ADE & R101 & - & - & - & - & 16.5 & 6.6 & - & 21.7 \\
X-Decoder~\cite{x-decoder} & ADE+ITP & Swin-T & 41.6 & 27.7 & 28.8 & 51.0 & - & - & - & - \\
OpenSeeD & ADE & Swin-T & 42.2 & 29.7 & 30.4 & 47.0 & 14.1 & 6 & 7.1 & 14.7 \\
\rowcolor{isabelline}
\textit{OpenSD} & ADE & Swin-T & \textbf{44.2} & \textbf{31} & \textbf{31.7} & \textbf{47.4} & \textbf{20.3} & \textbf{8.9} & \textbf{10.3} & \textbf{29.1} \\
\Xhline{0.8pt}
\end{tabular}
}
\caption{Open-vocabulary segmentation and detection from ADE to COCO.  The model is trained on ADE and tested on COCO and ADE. `ITP' means image-text pair data.
} 
\label{ade_to_coco}
\vspace{-1.5mm}
\end{table*}
\subsection{Results}
We present the results of our OpenSD model trained on COCO (or jointly with Object365) and ADE20K in Tab.~\ref{coco_to_ade}  and Tab.~\ref{ade_to_coco}, respectively. Though our OpenSD model is mainly designed for open-vocabulary segmentation and detection tasks, we compare it with existing methods on both closed- and open-vocabulary domains. We believe maintaining a competitive performance on both closed and open domains is important for providing a universal and omnipotent solution to segmentation and detection. 


\textbf{Closed-vocabulary Performance.} 
From Tab.~\ref{coco_to_ade}, it is evident that when only trained on COCO, our model demonstrates noticeable superiority over Mask2former on the closed-vocabulary COCO test set. Specifically, when using ResNet-50 as the backbone, OpenSD achieves 53.2 PQ, 45.2 mask AP, 48.8 box AP and 62.0 mIoU in panoptic segmentation, instance segmentation, object detection and semantic segmentation tasks, respectively, outperforming Mask2former by 1.3, 3.5, 4.3 and 0.3 points, respectively. When employing Swin-Tiny as the backbone, OpenSD achieves performance gains of 2.0, 4.2, 5.7, and 0.8 points over Mask2former in the four tasks. It is worth mentioning that for the Swin-L backbone, Mask2former trains the model for 100 epochs, while we only train OpenSD model for 50 epochs. Even though, OpenSD still outperforms Mask2former, particularly in instance segmentation and object detection. These results demonstrate the effectiveness of our proposed decoupled decoder learning approach.

Compared to other open-vocabulary methods, OpenSD also demonstrates clear advantages under the closed-vocabulary setting. Specifically, when using Swin-Tiny as the backbone and training on COCO, it outperforms OpenSeeD by 1.2, 1.9, 2.8, and 1.9 points respectively in the four tasks. Compared to X-Decoder, OpenSD achieves improvements of 2.6, 6.2, 8.2, and 1.6 points, and compared to SEEM, OpenSD outperforms it by 4.6, 8, and 2.8 points. The improvements are particularly prominent in instance segmentation and object detection tasks. These results can further validate that our decoupling decoder can mitigate the conflicts between thing categories and stuff categories, ensuring the effective learning of different tasks within a unified framework.
When  training jointly on COCO and Object365, our Swin-Tiny based model reaches 55.7, 48, 53.0 and 64.8 points in the four tasks, which still outperforms OpenSeeD by 0.5, 0.7, 1.1 and 1.1 points even with fewer training iterations.

Tab.~\ref{ade_to_coco} presents the closed-vocabulary results from ADE20K to COCO. We see that OpenSD exhibits significant performance advantages over other competing approaches in panoptic segmentation, instance segmentation and object detection tasks. For semantic segmentation, OpenSD achieves comparable performance to Mask2former and OpenSeeD, and lags slightly behind X-Decoder. However, it should be noted that X-Decoder benefits from the additional training with image-text pairs, which play an important role to improve its performance in semantic segmentation, as mentioned in \cite{openseed}.

\begin{table*}[!t]
\centering
\scriptsize
\scalebox{0.98}{
\setlength{\tabcolsep}{0.13mm}
\begin{tabular}{l|cc|ccccc ccccc ccccc ccccc ccccc}
\Xhline{0.8pt}
\rowcolor{LightCyan}
Method & Med. & Avg & \makecell{Air.-\\Par.} & Bottles & \makecell{Br.\\Tum.} & Chicken & Cows & \makecell{Ele.-\\Sha.} & Eleph. & Fruits & Gar. & \makecell{Gin.-\\Gar} & Hand & \makecell{Hand-\\Metal} & \makecell{House-\\Parts} & \makecell{HH.-\\Items} & \makecell{Nut.-\\Squi.} & Phones & Poles & Puppies & Rail & \makecell{Sal.\\Fil.} & Stra. & Tablets & Toolkits & Trash & W.M. \\
\hline\hline
X-Decoder(T) & 15.7 & 22.6 & 10.5 & 19 & 1.1 & 12 & 12 & 1.2 & 65.6 & 66.5 & 28.7 & 7.9 & 0.6 & 22.4 & 5.5 & 50.6 & 62.1 & 29.9 & 3.6 & 48.9 & 0.7 & 15 & 41.6 & 15.2 & 9.5 & 19.3 & 16.2 \\
OpenSeeD(T) & 21.5 & 33.9 & 12.2 & 27.4 & 5 & 68.7 & 21.5 & 0.3 & 73.3 & 72.9 & 7.3 & 6.2 & 92.4 & 62.3 & 0.5 & 55 & 63.6 & 2.4 & 4.6 & 63.8 & 5.4 & 15.6 & 85.3 & 32 & 4.8 & 14.5 & 51 \\
\rowcolor{isabelline}
\textit{OpenSD(T)} & \textbf{29.7} & \textbf{35.8} & 10.9 & 33.9 & 6.1 & 53.4 & 12.4 & 0.75 & 67.2 & 84.3 & 11.9 & 20.2 & 90 & 67 & 1.6 & 51.5 & 57.5 & 28.9 & 14.2 & 75.4 & 0.03 & 30.5 & 79.3 & 12.8 & 28.5 & 26 & 40.4 \\

\Xhline{0.8pt}
\end{tabular}
}
\caption{Zero-shot performance on the SeginW benchmark.
} 
\label{seginw}
\end{table*}
\begin{table*}[!t]
\centering
\scalebox{0.98}{
\setlength{\tabcolsep}{1.5mm}
\begin{tabular}{l|c|ccc|cccc|cccc}
\Xhline{0.8pt}
\rowcolor{LightCyan}
Method &  DD & \multicolumn{3}{c|}{Classifier} & \multicolumn{4}{c|}{COCO} &  \multicolumn{4}{c}{ADE}   \\
\hline\hline
 & & DP & In & Out & PQ & mask AP & box AP & mIoU & PQ & mask AP & box AP & mIoU  \\
\hline
M2F~\cite{mask2former} & \small{\XSolidBrush} & & fc & \small{\XSolidBrush} & 52.4 & 42.3 & 44.1 & 62.9 & - & - & - & - \\
M2F+LE & \small{\XSolidBrush} & \small{\XSolidBrush} & frozen & \small{\XSolidBrush} & 50.0 & 41.9 & 42.7 & 61.5 & 10.0 & 5.9 & 6.0 & 16.3 \\
M2F+LE & \small{\XSolidBrush} & \small{\XSolidBrush} & adapting & \small{\XSolidBrush} & 52.1 & 43.7 & 46.0 & 61.8 & 11.0 & 6.0 & 6.5 & 16.8 \\
M2F+LE & \small{\XSolidBrush} & \small{\Checkmark} & adapting & \small{\XSolidBrush} & 53.1 & 44.7 & 47.0 & 62.8 & 12.0 & 7.0 & 7.5 & 17.8 \\
M2F+LE & \small{\Checkmark} & \small{\Checkmark} & adapting & \small{\XSolidBrush} & 54.3 & 46.6 & 49.8 & 63.8  & 13.7 & 7.8 & 9.1 & 17.8 \\
M2F+LE+CLIP & \small{\Checkmark} & \small{\Checkmark} & adapting &  \small{\Checkmark} & 54.3 & 46.6 & 49.8 & 63.8 & 20.2 & 11.2 & 11.9 & 29.2 \\

\Xhline{0.8pt}
\end{tabular}
}
\caption{Ablation study for the road-map towards open-vocabulary segmentation and detection. All the models use Swin-T as the backbone and are trained on COCO for 36 epochs. `M2F' indicates Mask2former. `LE' indicates label embeddings. `DD' indicates decoupled decoder.  `DP' indicates decoupled prompts. 
} 
\label{ablation}
\vspace{-2.5mm}
\end{table*}
\textbf{Open-vocabulary Performance.}
The results of open-vocabulary experiments are also shown in Tab.~\ref{coco_to_ade}. We can see that OpenSD still achieves significant performance improvements over its competitors. On ADE20K, when using ResNet-50 as the backbone and jointly training on COCO and Object365, OpenSD achieves performance of 23, 14.3, 16.8, and 29.3 in panoptic segmentation, instance segmentation, object detection and semantic segmentation tasks, respectively. These results surpass the current state-of-the-art HIPIE by 4.6, 1.3, 0.6, and 2.5 points, respectively.
When Swin-Tiny is used as the backbone and COCO is used for training, OpenSD outperforms OpenSeeD by 3.6, 1.2, 1.5, and 10.1 points, and outperforms X-Decoder by 0.8, 1.3, and 3.7 points, respectively on the four tasks. When jointly training on COCO and Object365, OpenSD surpasses OpenSeeD by 2.4, 0.9, 0.8, and 7 points, respectively, although OpenSeeD employs additional self-training techniques and spends more iterations in training.

For the open-vocabulary performance on Cityscapes, OpenSD trained only on COCO achieves 38.2, 27.5, and 51 points in the four tasks,  surpassing OpenSeeD trained on COCO dataset by 2.5, 3.8, and 5.2 points, respectively. Similarly, when using the larger backbone Swin-Large, we consistently achieve performance improvements across all the four tasks, as shown in Tab.~\ref{coco_to_ade}. 

Tab.~\ref{ade_to_coco} compares the open-vocabulary performance from ADE to COCO.
We can see that OpenSD achieves 20.3, 8.9, 10.3 and 29.1 points, which are 6.2, 2.9, 3.2 and 14.4 points higher than OpenSeeD. 

\subsection{Segmentation in the Wild}
To further evaluate the generalization capability of our method, we conduct zero-shot testing on a diverse range of scenes using the Segmentation in the Wild (SeginW) benchmark \cite{x-decoder}, which consists of 25 segmentation datasets. The results are shown in Tab.~\ref{seginw}, from which we can see that OpenSD outperforms X-Decoder and OpenSeeD by 13.2 and 1.9 points, respectively, in terms of average precision. Moreover, in terms of median values, OpenSD exhibits even greater advantages, surpassing X-Decoder and OpenSeeD by 14 and 8.2 points, respectively.

\begin{table*}[!t]
\centering
\scalebox{0.95}{
\setlength{\tabcolsep}{1.8mm}
\begin{tabular}{c|ccccc}
\Xhline{0.8pt}
\rowcolor{LightCyan}
\diagbox{$\alpha$}{$\beta$} & 0.5 & 0.6 & 0.7 & 0.8 & 0.9  \\
\hline
\cellcolor{LightCyan}0.1 & 23.5,16.3,\textbf{12.8},11.3 & 25.0,17.4,\textbf{12.8},\textbf{11.4} & 26.6,19.0,12.4,11.1 & 28.0,19.7,11.7,10.5 & 28.8,18.6,10.9,10.\\
\cellcolor{LightCyan}0.2 & 25.6,16.5,12.7,11.2 & 26.7,18.1,12.7,11.3 & \textit{28.7,19.6,12.2,11.1} & 29.3,\textbf{20.2},11.6,10.5 & 29.3,18.8,10.8,9.9 \\
\cellcolor{LightCyan}0.3 & 27.4,16.5,12.5,11.1 & 28.4,18.0,12.5,11.1 & 29.5,19.4,12.1,10.8 & \textbf{29.7},10.1,11.4,10.3 & 29.5,19.0,10.6,9.7 \\
\cellcolor{LightCyan}0.4 & 28.1,15.9,12.2,10.8 & 29.1,17.4,12.2,10.8 & \textbf{29.7},18.8,11.8,10.6 & \textbf{29.7},19.8,11.1,10.0 & 29.5,18.3,10.4,9.4 \\
\Xhline{0.8pt}
\end{tabular}
}
\caption{Ensemble methods comparisons with zero-shot evaluation on ADE20K.
} 
\label{supp-ablation}
\end{table*}
\subsection{Ablation Study}

\textbf{Decoupled Prompt Learning.} We ablate each component of OpenSD in Tab.~\ref{ablation}. From the $2^{nd}$ row, we see that simply replacing the original classifier of Mask2former with label embeddings generated by a frozen text encoder leads to inferior performance. However, when prompt learning is applied to the text encoder (the $3^{rd}$ row), we observe performance improvements across various tasks, particularly in instance segmentation and object detection. This demonstrates that prompts have learned the ability to filter out low-quality masks and suppress repetitive predictions under the constraint of one-to-one label assignment loss.
Furthermore, when prompt learning is decoupled (the $4^{th}$ row), we observe further performance improvements. 

\textbf{Decoupled Decoder Learning.} From the $4^{th}$ and $5^{th}$ rows of Tab.~\ref{ablation}, we can observe consistent performance improvements in both COCO and ADE20K datasets by utilizing decoupled decoder learning. This indicates that the decoupled decoder can effectively mitigate the conflicts between different tasks, allowing each task to optimize its own performance independently. 

\textbf{Dual Classifiers.} From the $5^{th}$ row of Tab.~\ref{ablation}, one can observe that when only the in-vocabulary classifier is used, we achieve 13.7, 7.8, 9.1, and 17.8 points on the four tasks in ADE20K. However, by introducing the out-of-vocabulary classifier to handle unseen categories (the last row), we achieve performance improvements of 6.5, 3.4, 2.8, and 11.4 points on the four tasks. The improvements on panoptic segmentation and semantic segmentations are particularly great. This indicates that CLIP can significantly enhance the understanding of stuff categories in semantic and panoptic segmentations.

\textbf{Ensemble Strategy.} For in-vocabulary domain testing, we only use the in-vocabulary classifier and set both $\alpha$ and $\beta$ to 0. For out-of-vocabulary domain testing, we ablate the two hyper-parameters on ADE dataset with Swin-Tiny backbone in Tab.~\ref{supp-ablation}, where the four numbers in each cell represent the results of semantic segmentation, panoptic segmentation, object detection and instance segmentation, respectively.
As observed, the optimal results for different tasks are achieved at different values of $\alpha$ and $\beta$. For example, the best performance for semantic segmentation, with a score of 29.7, is achieved when ($\alpha$, $\beta$) is set to (0.4, 0.7), (0.3, 0.8), or (0.4, 0.8). The best performance for panoptic segmentation, with a score of 20.2, is obtained when $\alpha$ is set to 0.2 and $\beta$ is set to 0.8. The best performance for detection and instance segmentation is 12.8 and 11.4, respectively.

To strike a balance between the different tasks, we set $\alpha$ to 0.2 and $\beta$ to 0.7 by default, resulting in performance scores of 28.7 for semantic segmentation, 19.6 for panoptic segmentation, 12.2 for detection, and 11.1 for instance segmentation. These values provide a balanced trade-off among different tasks.

\begin{figure}[tbp]
    \centering
    \includegraphics[width=0.47\textwidth]{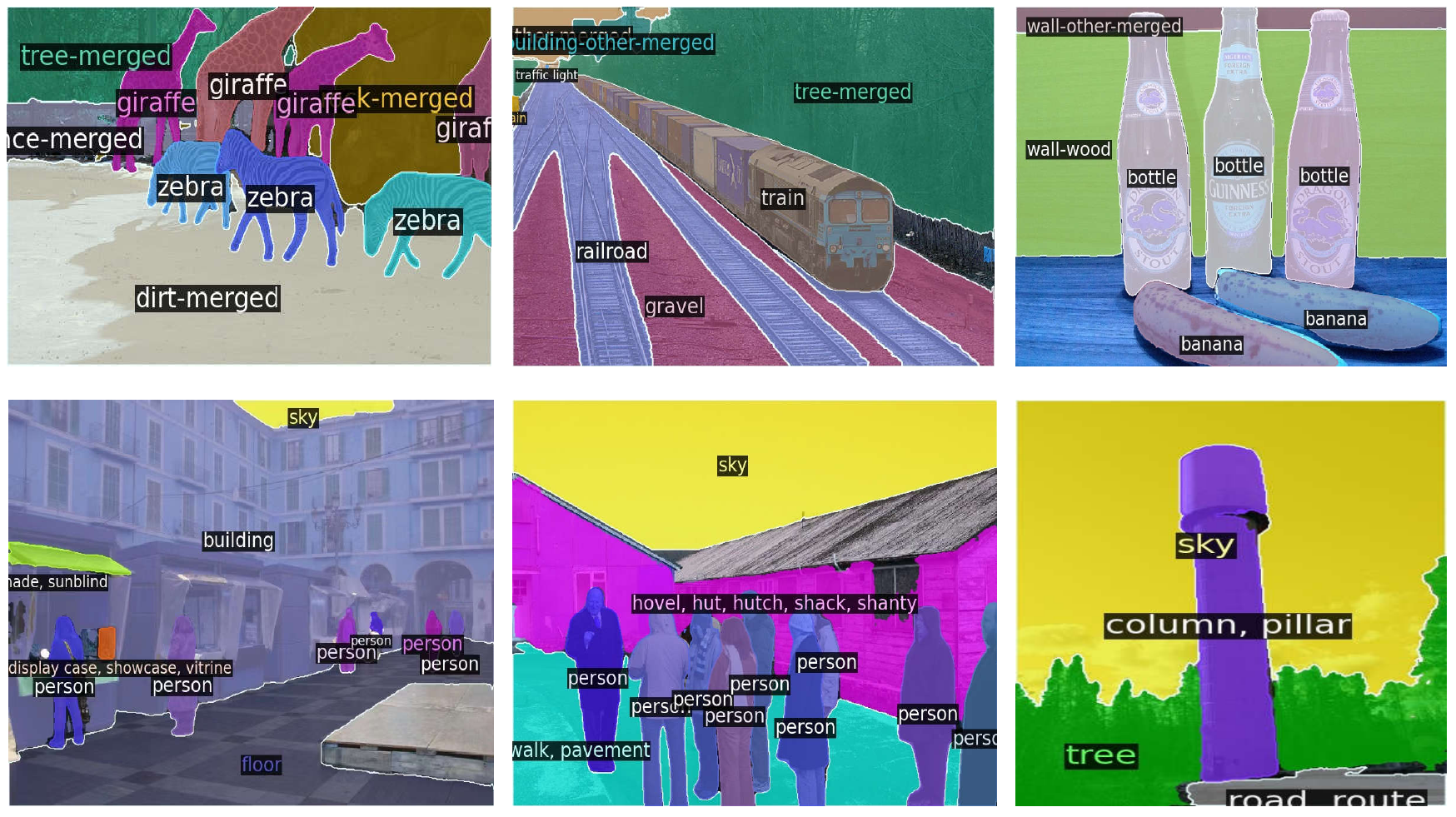}
    \vspace{-2mm}
    \caption{Visualization examples of OpenSD. The model is trained on COCO and tested on COCO closed-vocabulary domain (top) and ADE20K open-vocabulary domain (bottom).}
    \label{visualization}
    \vspace{-2mm}
\end{figure}

\begin{figure*}[tbp]
    \centering
    \includegraphics[height = 4.cm]{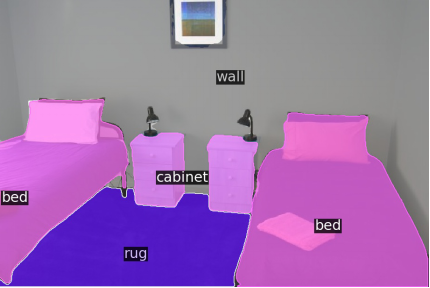}
    \includegraphics[height = 4.cm]{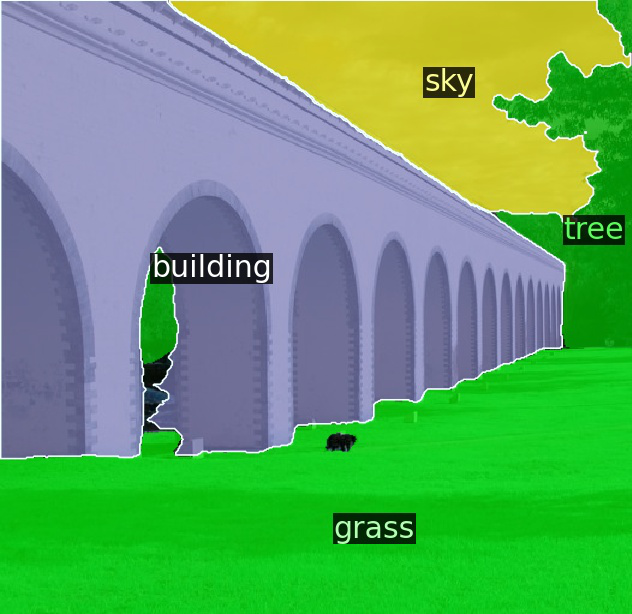}
    \includegraphics[height = 4.cm]{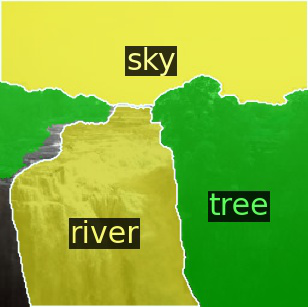}
    \includegraphics[height = 4.cm]{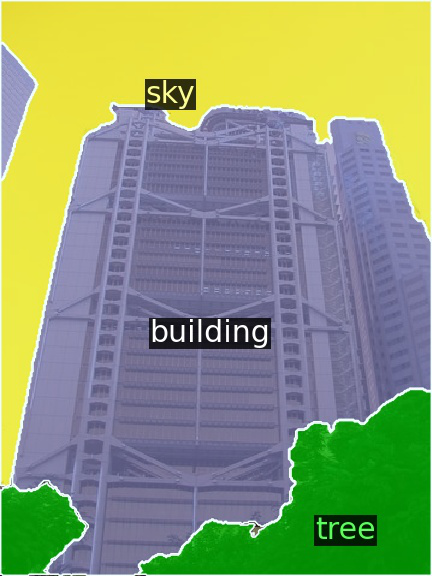}

    \subcaptionbox{\label{a}}{\includegraphics[height=4.cm]{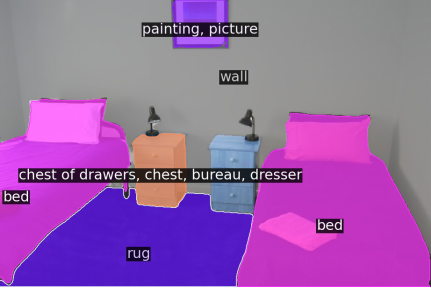}}
    \subcaptionbox{\label{b}}{\includegraphics[height=4.cm]{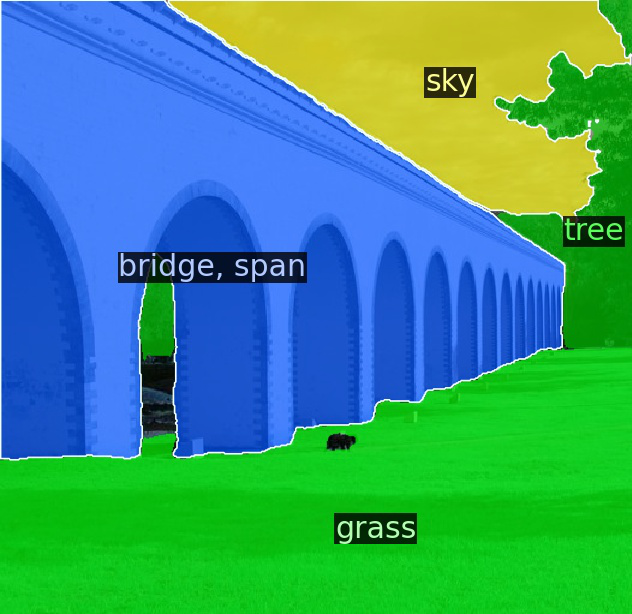}}
    \subcaptionbox{\label{c}}{\includegraphics[height=4.cm]{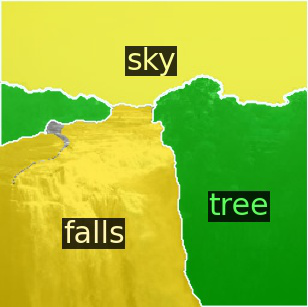}}
    \subcaptionbox{\label{d}}{\includegraphics[height=4.cm]{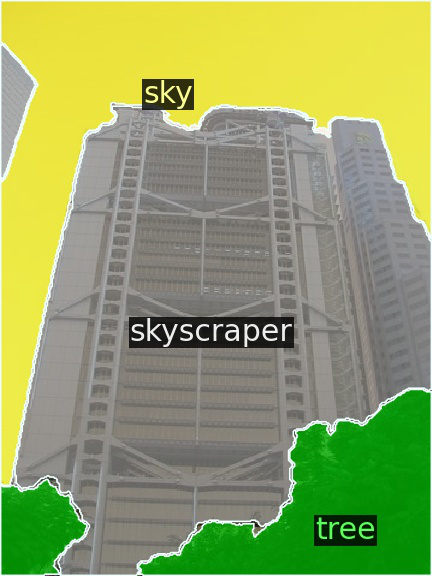}}

    \includegraphics[height = 3.31cm]{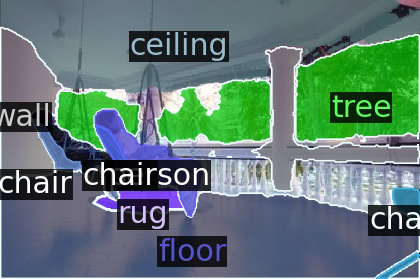}
    \includegraphics[height = 3.31cm]{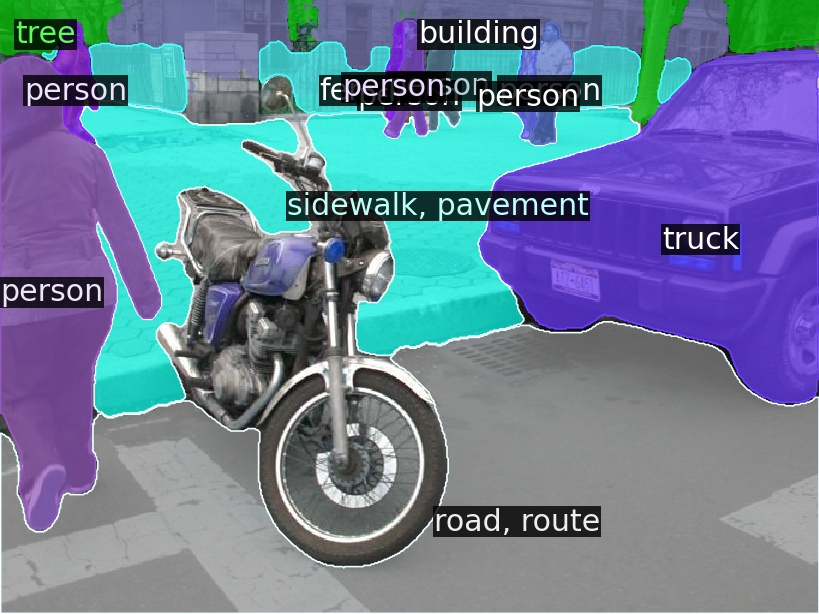}
    \includegraphics[height = 3.31cm]{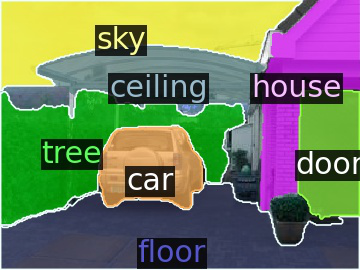}
    \includegraphics[height = 3.31cm]{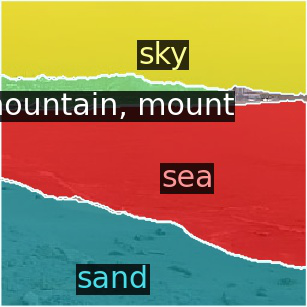}

    \subcaptionbox{\label{e}}{\includegraphics[height=3.31cm]{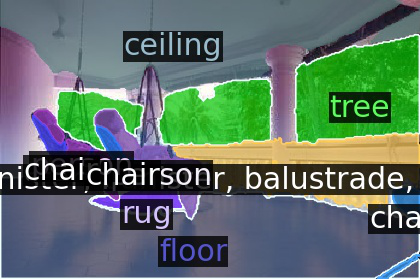}}
    \subcaptionbox{\label{f}}{\includegraphics[height=3.31cm]{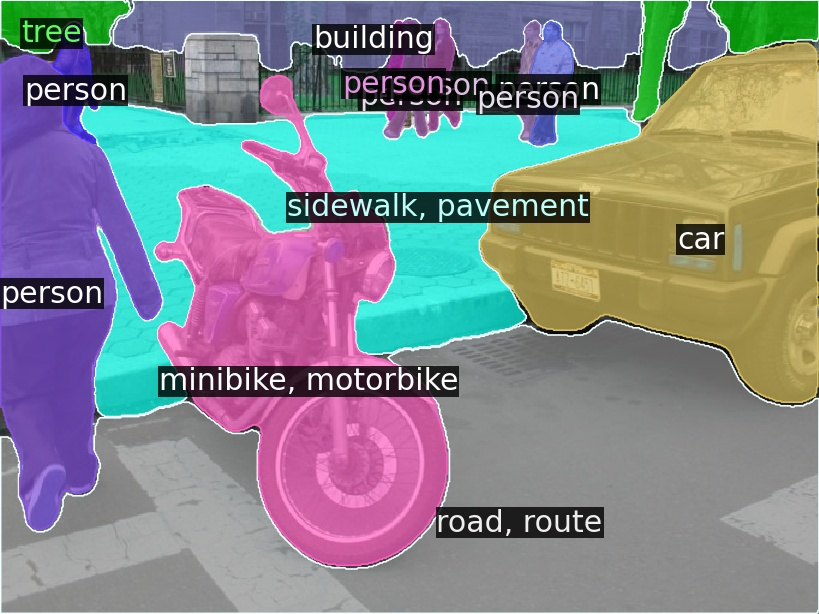}}
    \subcaptionbox{\label{g}}{\includegraphics[height=3.31cm]{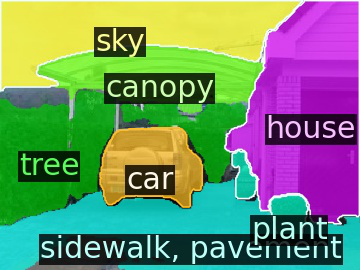}}
    \subcaptionbox{\label{h}}{\includegraphics[height=3.31cm]{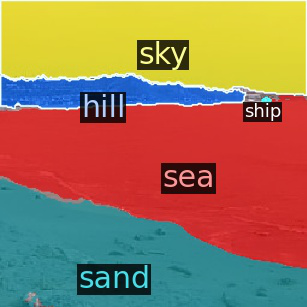}}
    \caption{Zero-shot inference examples on ADE20K. The top and bottom rows are the results of OpenSeeD and our method, respectively.}
    \label{supp-visualization}
\end{figure*}

\subsection{Qualitative Results}
In Fig.~\ref{visualization}, we visualize some segmentation examples of OpenSD. The top row showcases results in the closed-vocabulary domain on the COCO dataset. One can observe that OpenSD effectively distinguishes between different instance categories, such as zebra and giraffe in the first image, and accurately segments irregular-shaped stuff categories, such as the grave and railroad in the second image.

The bottom row of Fig.~\ref{visualization} displays the open-vocabulary segmentation results on the ADE20K dataset. OpenSD can accurately segment unseen categories, such as the sunblind category (first image) and the hovel category (second image) in semantic segmentation, and the column category (third image) in instance segmentation. These results highlight the exceptional generalization capability of OpenSD to arbitrary semantic categories.

We provide more visualization results of OpenSeeD and our OpenSD methods in Fig.~\ref{supp-visualization}.
We can have the following observations. Firstly, by leveraging CLIP to enhance the open-vocabulary capability, our OpenSD method effectively segments unseen categories, such as the bridge in Fig.~\ref{supp-visualization}\subref{b}, the falls in Fig.~\ref{supp-visualization}\subref{c}, the skyscraper in Fig.~\ref{supp-visualization}\subref{d}, the balustrade in Fig.~\ref{supp-visualization}\subref{e} and the canopy in Fig.~\ref{supp-visualization}\subref{g}. In contrast, OpenSeeD misclassifies these objects as seen categories. Secondly, by adapting CLIP's text encoder to be region-aware using prompt learning, we can differentiate between different instances. This is evident in the color-coded masks for different instances, such as the chest of drawers in Fig.~\ref{supp-visualization}\subref{a}. In contrast, OpenSeeD treats instances of the same category as a single object and assigns them the same color. Lastly, our proposed decoupled decoder learning strategy improves learning efficiency for different tasks, enabling more precise mask classification and segmentation. For example, OpenSD accurately segments the painting in Fig.~\ref{supp-visualization}\subref{a}, the motorbike in Fig.~\ref{supp-visualization}\subref{f} and the ship in Fig.~\ref{supp-visualization}\subref{h}, while OpenSeeD misses them. OpenSD also correctly identifies the pavement in Fig.~\ref{supp-visualization}\subref{g}, whereas OpenSeeD incorrectly classifies it as the floor.

\section{Conclusion}
In this paper, we presented OpenSD, a versatile framework designed to tackle the open-vocabulary segmentation and detection tasks using a unified architecture and shared parameters. Our approach leveraged decoupled decoder learning and prompted dual classifiers to effectively address the conflicts between different tasks and enhance the region-awareness of CLIP. Our experiments showed that OpenSD consistently outperformed existing methods in both closed- and open-vocabulary settings. With the Swin-Large backbone, OpenSD achieved improvements of 1.2, 2.4, 4.5, and 0.9 point in panoptic segmentation, instance segmentation, detection, and semantic segmentation tasks on COCO, respectively. By leveraging CLIP, OpenSD surpassed OpenSeeD by 6, 3.3, 3.8, and 11 points in the corresponding open-vocabulary tasks on ADE.
These new state-of-the-art results provided compelling evidence of the effectiveness and generalization capability of OpenSD in addressing various segmentation and detection tasks.

{
    \small
    \bibliographystyle{ieeenat_fullname}
    \bibliography{main}
}


\end{document}